\pdfoutput=1

\documentclass[11pt]{article}

\usepackage[final]{coling}

\usepackage{times}
\usepackage{latexsym}

\usepackage[T1]{fontenc}

\usepackage[utf8]{inputenc}

\usepackage{microtype}

\usepackage{inconsolata}

\usepackage{graphicx}

\usepackage{booktabs}
\usepackage{multirow}
\usepackage{bbm}

\usepackage{arydshln}
\makeatletter
\def\adl@drawiv#1#2#3{%
        \hskip.5\tabcolsep
        \xleaders#3{#2.5\@tempdimb #1{1}#2.5\@tempdimb}%
                #2\z@ plus1fil minus1fil\relax
        \hskip.5\tabcolsep}
\newcommand{\cdashlinelr}[1]{%
  \noalign{\vskip\aboverulesep
           \global\let\@dashdrawstore\adl@draw
           \global\let\adl@draw\adl@drawiv}
  \cdashline{#1}
  \noalign{\global\let\adl@draw\@dashdrawstore
           \vskip\belowrulesep}}
\makeatother

\title{Tougher Text, Smarter Models: \\Raising the Bar for Adversarial Defence Benchmarks}

\author{Yang Wang\textsuperscript{1,3} and Chenghua Lin\textsuperscript{1, 2 }\\
  \textsuperscript{1}Department of Computer Science, The University of Sheffield, UK \\
  \textsuperscript{2}Department of Computer Science, The University of Manchester, UK\\
  \textsuperscript{3}Automated Analytics, UK\\
  \texttt{y.wang4@sheffield.ac.uk} ~~~ \texttt{chenghua.lin@manchester.ac.uk}
}

\begin{document}
\maketitle
\begin{abstract}
Recent advancements in natural language processing have highlighted the vulnerability of deep learning models to adversarial attacks. 
While various defence mechanisms have been proposed, there is a lack of comprehensive benchmarks that evaluate these defences across diverse datasets, models, and tasks. 
In this work, we address this gap by presenting an extensive benchmark for textual adversarial defence that significantly expands upon previous work.
Our benchmark incorporates a wide range of datasets, evaluates state-of-the-art defence mechanisms, and extends the assessment to include critical tasks such as single-sentence classification, similarity and paraphrase identification, natural language inference, and commonsense reasoning. 
This work not only serves as a valuable resource for researchers and practitioners in the field of adversarial robustness but also identifies key areas for future research in textual adversarial defence. 
By establishing a new standard for benchmarking in this domain, we aim to accelerate progress towards more robust and reliable natural language processing systems.
\end{abstract}

\section{Introduction}

Recent advancements in natural language processing (NLP) have led to impressive performance on various tasks, but also exposed the vulnerability of deep learning models to adversarial attacks \cite{wang2021infobert,han2022text,wang2022ada,ranjan2023fooling, zeng2023certified, goyal2023survey, shayegani2023survey, huang2024survey}. 
While numerous defence mechanisms have been proposed to counter these threats, there is a lack of comprehensive benchmarks to evaluate their effectiveness across diverse settings.

The advent of adversarial training \cite{Goodfellow2014ExplainingAH} has demonstrated notable success in enhancing model robustness against small adversarial perturbations in computer vision. Traditional approaches adapt the training process to minimise empirical risk based on a \textit{robustness loss}, as opposed to the standard loss applied to clean input samples \cite{madry2018towards}. The robustness loss refers to the standard loss applied to the worst-case (i.e. loss-maximising) adversarial example for each training sample. 
In the context of NLP, however, adversarial training poses unique challenges due to the discrete nature of text. Specifically, the inner maximisation step required in the min-max formulation of adversarial training becomes computationally expensive \cite{yoo-qi-2021-towards-improving}. To address this, various methods have been proposed in the literature, ranging from augmenting the training set with adversarial examples tailored to a specific model \cite{si-etal-2021-better, dong2021towards, zhou-etal-2021-defense}, to more sophisticated optimisations in token-embedding space for the inner maximisation step \cite{Zhu2020FreeLB, li2021token, goyal2023survey}.

In parallel, other studies focus on structure-free regularisation methods for adversarial robustness. 
\citet{yang-etal-2023-domain} argue that encouraging higher entropy (i.e. uncertainty) in model outputs can enhance adversarial robustness. 
They emphasise the need to understand the \textit{inherent robustness} properties of models, focusing on those that are flexible, simple, and not overly specialised for specific types of text adversarial attacks, as well as the interplay between a model's confidence and robustness. 
Building on this idea, they highlight that entropy regularisation techniques, such as label smoothing \cite{szegedy2015rethinking, szegedy2016rethinking}, can implicitly contribute to adversarial robustness by addressing model overconfidence. 
Similarly, \citet{raina-etal-2024-extreme} proposed training-time temperature scaling as a defence mechanism. 
They empirically demonstrated that highly miscalibrated models \cite{guo2017calibration} interfere with an adversarial attacker's ability to find meaningful search directions due to the little sensitivity in the predicted probabilities. 

Unlike those adversarial training-based methods, which rely on manipulating the token-embedding space for inner maximisation to enhance adversarial robustness, regularisation-based approaches offer a more attractive and effective alternative. 
These regularisation-based methods are synonyms-agnostic and structure-free, which can be seamlessly applied across a broad spectrum of NLP tasks, extending beyond traditional text classification. 
To the best of our knowledge, the most recent benchmark in this area was established by \citet{li-etal-2021-searching}. 
They offered foundational insights but limited their focus to text classification tasks, evaluating only two datasets with defence methods developed prior to 2021. 
In contrast, our work broadens the evaluation by emphasising synonyms-agnostic and structure-free methods, ensuring broad applicability and relevance to a wider array of NLP challenges and tasks.
Our contributions include:

\begin{enumerate}
    \item We argue that the existing adversarial defence benchmark, as established by \citet{li-etal-2021-searching}, is limited in scope. In response, we extend the evaluation to include more NLP datasets, tasks, models, and recent advanced adversarial defence techniques.
    \item We propose TTSO++, a variant of training-time temperature scaling that incorporates dynamic confidence adjustment through an entropy term. This adaptation enhances robustness against adversarial attacks, especially under TextFooler and TextBugger scenarios.
\end{enumerate}
Our code  is available at \url{https://github.com/PuReDefence/AdvBench4Text}.

\section{Background}

The vulnerability of deep learning models to adversarial attacks has become a significant concern in NLP. This section provides an overview of adversarial attacks and defences in NLP, with a particular focus on flexible defence methods that can be adapted to various NLP tasks.

\subsection{Adversarial Attacks}

Adversarial attacks in NLP aim to manipulate input text in ways that preserve semantic meaning but cause model misclassification. 
Following notation in \citet{raina2023sample} the distance between the benign sample $x$ and the adversarial example $\tilde{x}$ can be measured via a proxy function $\mathcal{G}(x, \tilde{x}) \leq \epsilon$, where $\epsilon$ represents the maximum imperceptibility threshold. 
\citet{goyal2023survey} categorised these attacks based on the attacker's knowledge (white-box vs. black-box), the perturbation level (character, word, or sentence-level), and the attack goal (targeted vs. untargeted). 
Common attack methods include word substitution \cite{ren2019generating, zang-etal-2020-word, li-etal-2020-bert-attack, Garg2020BAEBA, jin2020bert, maheshwary2021generating, waghela2024modified, lu2024less}, character manipulation \cite{gao2018black, eger-etal-2019-text, eger2019text, pruthi-etal-2019-combating, liu-etal-2022-character, rocamora2024revisiting}, and sentence paraphrasing \cite{ribeiro2018semantically, iyyer-etal-2018-adversarial, zhao2018generating, li2020contextualized, li-etal-2021-contextualized}. 
Many of these popular attack methods are implemented in the TextAttack library \cite{morris2020textattack}.

\subsection{Adversarial Defences}

In this section, we will discuss two different types of adversarial defence methods.

\subsubsection{Adversarial Training-based Methods}


Numerous defence methods have been proposed to counter adversarial threats. 
In computer vision, adversarial training \cite{goodfellow2014explaining} minimises empirical risk from worst-case adversarial examples, but its inner maximisation step is computationally expensive for NLP models. 
To address this, a group of adversarial training methods like PGD \cite{madry2018towards}, FreeLB \cite{Zhu2020FreeLB}, and TAVAT \cite{li2021token} accelerate optimisation by identifying adversarial examples in the token-embedding space. 

Despite their efficiency, the limited success of these methods is often attributed to perturbations in the embedding space, which may not adequately represent true adversarial examples in natural language. 
To mitigate this issue, approaches such as ASCC \cite{dong2021towards} and DNE \cite{zhou2021defense} proposed a more meaningful embedding perturbation space, defining it as the convex hull of word synonyms. 
While these methods offer improved robustness, they require pre-computation of synonyms, limiting their adaptability and effectiveness against diverse adversarial attacks. 
In light of these challenges, we emphasise the need for synonyms-agnostic and structure-free defence strategies, which provide broader applicability across NLP tasks. 
In practical scenarios, defenders should not rely on prior knowledge of the adversary's mechanisms for generating synonyms, as this can limit the robustness of the defence.

\subsubsection{Regularisation-based Methods}

Regularisation-based methods have emerged as a more flexible and generalisable approach to adversarial defence in NLP, particularly because they do not rely on model structures or synonym sets, making them adaptable across a wide range of tasks. 
Methods such as Flooding-X \cite{liu-etal-2022-flooding}, adversarial label smoothing \cite{yang-etal-2023-domain}, and temperature scaling \cite{raina-etal-2024-extreme} have demonstrated notable effectiveness in enhancing adversarial robustness. 

Flooding-X \cite{liu-etal-2022-flooding} aims to prevent overconfidence in model predictions by maintaining the loss around a pre-defined ``flood'' level \cite{10.5555/3524938.3525366}, thereby mitigating the model's susceptibility to adversarial perturbations. 
Label smoothing \cite{szegedy2016rethinking}, on the other hand, modifies the training objective by softening the hard labels, distributing a small amount of probability mass across all classes, which helps in reducing the model's confidence in incorrect predictions. \citet{yang-etal-2023-domain} extensively studied standard label smoothing and its adversarial variant \cite{ren2022adversarial}, and showed that label smoothing can improve robustness to textual adversarial attacks (both black-box and white-box) and mitigate overconfident errors on adversarial examples. 
Additionally, \citet{raina-etal-2024-extreme} highlighted that the extreme class confidence exhibited by miscalibrated models \cite{guo2017calibration} creates an illusion of robustness (IOR). 
To address this, they proposed training-time temperature scaling as a defence mechanism to improve \textit{true} robustness against unseen attacks. 
Their empirical results showed that highly miscalibrated models impede adversarial attackers by reducing sensitivity in predicted probabilities, thereby limiting the attacker's ability to identify meaningful search directions.

Together, these regularisation-based methods provide a synonyms-agnostic and structure-free framework for adversarial defence, making them well-suited for diverse NLP tasks without requiring prior knowledge of adversarial strategies.

\section{Experiments}

\subsection{Datasets}

\begin{table}[!tp]\centering
\resizebox{\linewidth}{!}{
\begin{tabular}{lrrrrr}\toprule
\textbf{Dataset} &\textbf{\# Classes} &\textbf{Train} &\textbf{Validation} &\textbf{Test} &\textbf{Task} \\\midrule
SST2 &2 &6920 &872 &1821 &single-sentence classification \\
MR &2 &8530 &1066 &1066 &single-sentence classification \\
MRPC &2 &3668 &408 &1725 &paraphrase identification \\
SciTail &2 &23088 &2126 &1304 &natural language inference \\
SIQA &3 &33410 &1954 &- &commonsense reasoning \\
CSQA &5 &9741 &1221 &- &commonsense reasoning \\
\bottomrule
\end{tabular}
}
\caption{Dataset statistics.}\label{tab:statistics}
\end{table}

Experiments are carried out on six NLP datasets (statistics summarised in Table~\ref{tab:statistics}), including different tasks: single-sentence classification, similarity and paraphrase identification, natural language inference, and commonsense reasoning. 

SST2 \cite{socher2013recursive} is a binary sentiment classification task where each sample consists of a single sentence from movie reviews. The objective is to predict whether a given sentence expresses positive or negative sentiment. MR \cite{Pang+Lee:05a} is another binary sentiment classification dataset similar to SST-2, based on movie reviews. 
Each sentence is labelled as expressing either positive or negative sentiment. MRPC \cite{dolan2005automatically} is a binary classification dataset for similarity and paraphrase identification, where the task is to determine whether two sentences in a pair are semantically equivalent. SciTail \cite{khot2018scitail} is a natural language inference (NLI) dataset designed to test a model's ability to recognise entailment. SIQA \cite{sap2019socialiqa} is a commonsense reasoning dataset where the goal is to choose the most appropriate answer from three options to questions about everyday social situations. 
SIQA presents a challenge in understanding social dynamics and reasoning beyond explicit facts. 
CSQA \cite{talmor-etal-2019-commonsenseqa} is another multiple-choice question answering dataset that requires different types of commonsense knowledge to predict the correct answers.

These datasets cover a range of tasks, including single-sentence classification, similarity and paraphrase identification, natural language inference, and commonsense reasoning, enabling comprehensive evaluation across multiple dimensions of language understanding. Each dataset was carefully selected to ensure diversity in task complexity and linguistic phenomena, providing a robust benchmark for assessing model performance in various natural language understanding (NLU) tasks.


\subsection{Models}

\begin{table}[!t]\centering
\resizebox{\linewidth}{!}{
\begin{tabular}{ccr}\toprule
\textbf{Model} &\textbf{Checkpoint} &\textbf{Params} \\\midrule
BERT-base &google-bert/bert-base-uncased &109M \\
RoBERTa-base &FacebookAI/roberta-base &124M \\
DeBERTa-base &microsoft/deberta-v3-base &184M \\
BGE-M3 &BAAI/bge-m3 &567M \\
\bottomrule
\end{tabular}
}
\caption{\label{tab:model_card}
Pre-trained language models (PLMs) checkpoints from HuggingFace Hub. \textbf{Model}: Lists the names of different PLMs. \textbf{Checkpoint}: Specifies the HuggingFace checkpoint name with each model. \textbf{Params}: Indicates the number of parameters in each model.}
\end{table}

We follow existing adversarial robustness literature \cite{raina-etal-2024-extreme, zhao2024disentangled, b35be69309694ab1b32513fe4e679660} and use Transformer \cite{vaswani2017attention} encoders, which are state-of-the-art on many NLP tasks\footnote{Appendix~\ref{appendix:generative_llms} shows the performance of encoder-only models relative to generative LLMs for many classification tasks.}. Specifically, we consider the base variants of BERT \cite{devlin-etal-2019-bert}, RoBERTa \cite{Liu2019RoBERTaAR}, and DeBERTa \cite{He2020DeBERTaDB}, which are the most commonly used baseline models in prior adversarial defence studies. 
To extend this evaluation, we also assess adversarial robustness using a more recent state-of-the-art embedding model BGE-M3 \cite{chen-etal-2024-m3}. 
A summary of all evaluated models is presented in Table~\ref{tab:model_card}. 

While generative large language models (LLMs) such as Llama \cite{dubey2024llama} and ChatGPT \cite{chatgpt} have demonstrated impressive capabilities in various NLP tasks, their inclusion in adversarial robustness evaluations for our benchmark datasets 
is not appropriate. 
Our preliminary experiments (summarised in Table~\ref{tab:generative_llms}) on some of our benchmark datasets show that generative LLMs like Llama3-8B \cite{dubey2024llama} and Phi3-3.8B \cite{abdin2024phi} perform poorly on clean accuracy compared to smaller, discriminative models such as BERT, RoBERTa, and DeBERTa. 
Despite their large parameter counts, these models consistently underperform on clean (without attack) classification tasks, which undermines the significance of their adversarial robustness, as robustness should be evaluated in the context of maintaining high accuracy on before-attack data. 
Given the high computational cost and lower clean accuracy of these models, it is misleading to report a high after-attack accuracy and a low attack success rate without considering their poor baseline before-attack performance.


\subsection{Adversarial Defence Approaches}
\label{section:adversarial_defence}

We consider seven defence baselines in our benchmark: PGD \cite{madry2018towards}, FreeLB \cite{Zhu2020FreeLB}, TAVAT \cite{li2021token}, Flooding-X \cite{liu-etal-2022-flooding}, standard label smoothing (SLS) and adversarial label smoothing (ALS) \cite{yang-etal-2023-domain}, and training-time temperature scaling optimisation (TTSO) \cite{raina-etal-2024-extreme}. 

We further create a simple variant of the baseline TTSO that uses entropy-based temperature scaling during training, named TTSO++. This approach adjusts the temperature based on the entropy of the prediction distribution. High entropy indicates that the model is uncertain, so a lower temperature can be applied to sharpen the distribution. Conversely, low entropy (high certainty) can be smoothed by applying a higher temperature. The temperature $T$ is adjusted according to the entropy $H(\cdot)$ of the softmax distribution $p$:

\begin{equation}
    T = T_{base} + \alpha \cdot H(p)
\end{equation}
where $H(p)$ is the entropy of the softmax probabilities $p$, $T_{base}$ is the base temperature, and $\alpha$ is a scaling factor controlling how strongly the temperature reacts to uncertainty. 
By adding entropy $H(p)$ to the temperature scaling formula, we introduce \textit{dynamic confidence adjustment} based on the model's uncertainty. 
Note that \citet{10.1007/s00521-024-09505-4} also proposed an entropy-based temperature scaling method, but we introduce a simpler one that does not need any learnable parameters.


\subsection{Evaluation Metrics}

We follow the conventions in the literature \cite{li-etal-2021-searching, liu-etal-2022-flooding, lee2022query} to evaluate our benchmark. We leverage TextFooler \cite{jin2020bert} and TextBugger \cite{Li2018TextBuggerGA} to attack the victim models and measure the empirical performance. Both attackers are implemented using the default settings from the TextAttack library \cite{morris2020textattack}. 
While we acknowledge the advancements in attack techniques, TextAttack currently provides limited support for newer methods and only includes adversarial attack methods developed prior to 2021. 
Similarly, another widely-used OpenAttack \cite{zeng-etal-2021-openattack} library only covers adversarial attack methods up to 2020. 
Therefore, we focused on three well-established, general-purpose attack methods that are widely recognised for evaluating adversarial robustness \cite{wang2022rethinking, yang-etal-2023-fantastic, zhan-etal-2023-similarizing, hu-etal-2023-mask, yang-etal-2023-domain, lu2024less, ji-etal-2024-advancing, zhang-etal-2024-random, zhao2024disentangled}. 

To quantify the impact of each adversarial attack, we follow prior works \cite{li-etal-2021-searching, liu-etal-2022-flooding, hu-etal-2023-mask} and report the following metrics: accuracy under attack (\textsc{Aua}), attack success rate (\textsc{Asr}), and the average number of queries (\textsc{AvgQ}) required to successfully attack a model. Additionally, we provide the before-attack accuracy to offer a baseline for comparison, and quantify the relative performance decline using performance drop rate (\textsc{Pdr}).

\noindent\textbf{Clean Accuracy (\textsc{Acc})}~~measures the accuracy of the model on the before-attack dataset. It provides a baseline for how well the model performs without adversarial interference.

\noindent\textbf{Accuracy Under Attack (\textsc{Aua})}~~evaluates the accuracy of the model when subjected to adversarial examples. A higher \textsc{Aua} indicates better robustness against adversarial attacks.

\noindent\textbf{Attack Success Rate (\textsc{Asr})}~~is the percentage of adversarial attacks that successfully cause the model to misclassify. A lower \textsc{Asr} signifies a more robust model.

\noindent\textbf{Number of Queries (\textsc{AvgQ})}~~quantifies the average number of queries made to the model by an adversarial attack to achieve success. A higher number implies the model is harder to attack \cite{li2021searching}. 

\noindent\textbf{Performance Drop Rate (\textsc{Pdr})}~~quantifies the relative performance decline, and provides a normalised measure for comparing different attacks \cite{zhu2023promptbench}. 
\textsc{Apdr} stands for average \textsc{Pdr} across different attacks.

In contrast to prior work \cite{dong2021towards, bao2021defending, zheng-etal-2022-plugat, liu2022flooding, wang2022rethinking, hu-etal-2023-mask, zeng2023certified, zhan-etal-2023-similarizing, wang2023rmlm}, which often limits evaluations to a small subset of test samples from their datasets, we advocate for the inclusion of the \textit{entire} test set across all datasets. 
This comprehensive evaluation ensures a more robust assessment of the defence methods' effectiveness. 
Such an approach contrasts with the prevailing practice in the field, where evaluations may be restricted to a small portion of the available test data, potentially leading to an incomplete representation of a model's performance across diverse scenarios.

\subsection{Implementation Details}

All experiments were conducted using the HuggingFace framework \cite{wolf2020transformers} to leverage pre-trained model weights. 
For the adversarial training-based methods, including PGD, FreeLB, and TAVAT, we used the default hyper-parameters provided by the TextDefender library \cite{li-etal-2021-searching}. 
The default hyper-parameters for each adversarial training baseline are:  adversarial iterations of 5, adversarial learning rate of 0.03, adversarial initialisation magnitude of 0.05, adversarial maximum norm of 1, adversarial norm type of $l_2$. 
For experiments involving SLS and ALS, we performed a hyper-parameter search for the label smoothing coefficient from the set $\{ 0.1, 0.2, 0.3, 0.4, 0.5 \}$. 
In experiments involving TTSO and TTSO++, we applied the same high temperature $T=10$ to every instance and scaling factor $\alpha = 0.5$ by default. 
The learning rate was optimised by selecting the model that achieved the highest validation accuracy after fine-tuning for four epochs, with candidate values for the learning rate drawn from the set $\{ 1e-5, 2e-5, 5e-5 \}$. 
For commonsense reasoning datasets, we follow \citet{branco2021shortcutted} to fine-tune the pre-trained model, converting the multiple-choice task into a sequence-ranking problem, as outlined in \citet{liu2019multi}. We process the elements of input pairs separately, generating a score for each, with the maximum score corresponding to the selected answer. 
All experiments were executed on a single NVIDIA RTX 4090 GPU with 24GB of memory.

\subsection{Results}

\subsubsection{Robustness in Classification Tasks}

\begin{table*}[!t]\centering
\scriptsize
\begin{tabular}{lrrrrrrrrrrr}\toprule
\multirow{3}{*}{\textbf{Dataset}} &\multirow{3}{*}{\textbf{Defence}} &\multicolumn{3}{c}{\textbf{BERT}} &\multicolumn{3}{c}{\textbf{RoBERTa}} &\multicolumn{3}{c}{\textbf{DeBERTa}} \\\cmidrule{3-11}
& &\multirow{2}{*}{\textbf{\textsc{Acc}$\uparrow$}} &\multicolumn{2}{c}{\textbf{\textsc{Aua}$\uparrow$}} &\multirow{2}{*}{\textbf{\textsc{Acc}$\uparrow$}} &\multicolumn{2}{c}{\textbf{\textsc{Aua}$\uparrow$}} &\multirow{2}{*}{\textbf{\textsc{Acc}$\uparrow$}} &\multicolumn{2}{c}{\textbf{\textsc{Aua}$\uparrow$}} \\\cmidrule{4-5}\cmidrule{7-8}\cmidrule{10-11}
& & &\textbf{TF} &\textbf{TB} & &\textbf{TF} &\textbf{TB} & &\textbf{TF} &\textbf{TB} \\\midrule
\multirow{9}{*}{SST2} 
&- &91.54 &6.59 &28.08 &93.79 &7.80 &31.52 &95.22 &8.57 &39.65 \\
&PGD &\textbf{92.64} &8.73 &34.27 &94.29 &8.79 &34.21 &94.67 &8.73 &38.88 \\
&FreeLB &91.98 &8.57 &31.80 &95.11 &6.43 &36.90 &95.22 &10.32 &44.10 \\
&TAVAT &\textbf{92.64} &10.71 &34.32 &\textbf{95.17} &9.23 &37.40 &\textbf{95.83} &11.04 &47.01 \\
&Flooding-X &89.84 &11.64 &30.37 &94.95 &5.44 &29.65 &95.22 &7.36 &31.74 \\
&SLS &91.76 &12.25 &39.21 &94.34 &11.64 &44.81 &95.22 &22.24 &54.48 \\
&ALS &91.21 &15.54 &39.37 &94.51 &23.50 &52.50 &95.55 &15.76 &52.22 \\
&TTSO &91.71 &41.63 &50.85 &94.67 &45.63 &56.84 &95.66 &55.02 &65.84 \\
\cdashlinelr{2-11}
&TTSO++ &91.76 &\textbf{43.27} &\textbf{53.27} &94.78 &\textbf{48.93} &\textbf{59.91} &95.55 &\textbf{56.07} &\textbf{66.23} \\

\midrule
\multirow{9}{*}{MR} 
&- &85.55 &3.94 &22.80 &88.65 &6.75 &31.80 &90.71 &9.94 &34.24 \\
&PGD &85.93 &12.85 &35.46 &87.90 &6.29 &34.33 &89.21 &5.35 &30.11 \\
&FreeLB &86.30 &6.66 &28.71 &\textbf{89.12} &5.53 &33.58 &91.18 &8.63 &34.80 \\
&TAVAT &86.02 &8.26 &30.11 &87.99 &6.19 &32.46 &90.90 &10.60 &38.37 \\
&Flooding-X &85.55 &5.44 &27.02 &88.74 &7.79 &31.05 &\textbf{91.37} &6.66 &35.46 \\
&SLS &\textbf{86.59} &13.60 &36.49 &88.09 &20.83 &42.59 &90.62 &17.45 &45.31 \\
&ALS &85.83 &11.07 &33.77 &87.90 &17.07 &43.34 &89.96 &17.92 &46.53 \\
&TTSO &86.02 &35.27 &43.06 &87.71 &42.40 &51.97 &90.34 &47.84 &56.75 \\
\cdashlinelr{2-11}
&TTSO++ &86.02 &\textbf{38.09} &\textbf{45.31} &87.62 &\textbf{43.39} &\textbf{52.94} &90.24 &\textbf{49.44} &\textbf{57.22} \\

\midrule
\multirow{9}{*}{MRPC} 
&- &84.40 &2.32 &3.25 &86.96 &6.09 &9.57 &87.94 &2.96 &9.86 \\
&PGD &84.06 &9.86 &11.25 &87.48 &5.28 &11.54 &87.83 &3.71 &10.55 \\
&FreeLB &\textbf{85.45} &11.48 &11.65 &87.54 &6.38 &11.71 &86.43 &8.58 &15.07 \\
&TAVAT &84.29 &8.70 &10.43 &\textbf{87.59} &9.97 &15.07 &88.06 &6.61 &16.23 \\
&Flooding-X &82.67 &7.48 &7.88 &87.48 &4.75 &8.41 &88.35 &3.30 &10.20 \\
&SLS &84.52 &6.32 &7.36 &86.84 &9.22 &12.35 &\textbf{88.46} &8.58 &12.87 \\
&ALS &82.96 &5.97 &7.83 &86.03 &10.61 &13.28 &\textbf{88.46} &6.14 &17.45 \\
&TTSO &83.77 &41.62 &39.71 &86.78 &46.38 &41.91 &87.54 &50.78 &54.38 \\
\cdashlinelr{2-11}
&TTSO++ &83.48 &\textbf{42.49} &\textbf{40.92} &86.84 &\textbf{50.43} &\textbf{42.90} &87.59 &\textbf{50.99} &\textbf{55.12} \\

\midrule
\multirow{9}{*}{SciTail} 
&- &92.80 &44.45 &32.22 &93.60 &42.80 &31.70 &95.53 &47.04 &33.82 \\
&PGD &93.09 &50.19 &32.69 &93.09 &43.27 &31.42 &94.36 &43.32 &30.39 \\
&FreeLB &93.60 &47.04 &32.60 &\textbf{93.79} &44.21 &31.51 &95.67 &47.04 &33.68 \\
&TAVAT &92.29 &52.21 &30.48 &\textbf{93.79} &46.38 &34.71 &\textbf{96.52} &50.19 &39.98 \\
&Flooding-X &91.58 &49.62 &35.28 &92.43 &43.32 &29.02 &94.73 &46.28 &32.22 \\
&SLS &92.33 &48.02 &34.85 &93.60 &45.58 &35.79 &95.67 &48.64 &35.04 \\
&ALS &92.57 &50.80 &33.44 &92.90 &44.17 &34.48 &95.16 &50.85 &43.09 \\
&TTSO &92.33 &52.02 &48.21 &92.52 &51.27 &47.22 &95.63 &55.13 &50.05 \\
\cdashlinelr{2-11}
&TTSO++ &\textbf{93.74} &\textbf{54.47} &\textbf{49.86} &93.41 &\textbf{53.23} &\textbf{49.12} &95.25 &\textbf{56.02} &\textbf{51.32} \\
\bottomrule
\end{tabular}
\caption{The experiment results of different defence methods. TF: TextFooler. TB: TextBugger. The best performance is marked in \textbf{bold}.}\label{tab:main_v1}
\end{table*}

Table~\ref{tab:main_v1} presents the experimental results trained with various defence methods. 
Notably, TTSO and TTSO++ consistently outperform other baselines, achieving superior \textsc{Aua} across diverse attacks (TextFooler and TextBugger) and model architectures (BERT, RoBERTa, and DeBERTa). 
This robustness can be attributed to their ability to counteract the Illusion of Robustness (IOR) by addressing model miscalibration \cite{raina-etal-2024-extreme}, a key factor behind overconfidence in adversarial scenarios. 
Unlike token-level embedding perturbation techniques such as PGD, FreeLB, and TAVAT, which often lead to overfitting specific attack patterns without enhancing overall model uncertainty, TTSO and TTSO++ effectively recalibrate model confidence by softening predictions, setting a new benchmark for adversarial defence strategies. 

In comparison, methods like SLS and ALS emerge as flexible and lightweight alternatives to adversarial training-based methods. 
While approaches such as PGD, FreeLB, or TAVAT require computationally expensive inner maximisation steps during training and sometimes degrade performance under adversarial conditions, SLS and ALS offer significant improvements in adversarial robustness with minimal additional complexity. 
As shown in Table~\ref{tab:main_v1}, Flooding-X consistently underperforms compared to other baselines.
This poor performance aligns with the findings of \citet{zhu2023exploring}, who found flooding techniques ineffective for adversarial robustness. 
By maintaining the loss above a threshold, we argue that Flooding-X will hinder the model's ability to minimise adversarial loss and learn intricate decision boundaries. 
Its non-targeted regularisation treats all examples uniformly, lacking the specificity needed to counter adversarial attacks. 
While aimed at improving generalisation, Flooding-X appears to compromise the nuanced feature learning required for robust adversarial performance.

\subsubsection{Evaluate on Embedding-based Model}

\begin{table*}[!t]\centering
\scriptsize
\begin{tabular}{lrrrrrrrrrr}\toprule
\multirow{2}{*}{\textbf{Dataset}} &\multirow{2}{*}{\textbf{Defence}} &\multirow{2}{*}{\textbf{\textsc{Acc}$\uparrow$}} &\multicolumn{3}{c}{\textbf{TextFooler}} &\multicolumn{3}{c}{\textbf{TextBugger}} &\multirow{2}{*}{\textbf{\textsc{Apdr}$\downarrow$}} \\\cmidrule{4-9}
& & &\textbf{\textsc{Aua}$\uparrow$} &\textbf{\textsc{Asr}$\downarrow$} &\textbf{\textsc{AvgQ}$\uparrow$} &\textbf{\textsc{Aua}$\uparrow$} &\textbf{\textsc{Asr}$\downarrow$} &\textbf{\textsc{AvgQ}$\uparrow$} & \\\midrule
\multirow{9}{*}{SST2} &- &93.36 &6.59 &92.94 &91.14 &38.00 &59.29 &43.33 &76.11 \\
&PGD &92.86 &6.59 &92.90 &94.72 &40.14 &56.77 &44.30 &74.84 \\
&FreeLB &93.85 &6.43 &93.15 &90.58 &39.43 &57.99 &43.79 &75.57 \\
&TAVAT &\textbf{94.89} &6.70 &92.94 &92.85 &40.47 &57.35 &44.15 &75.14 \\
&Flooding-X &93.03 &4.50 &95.16 &85.47 &34.43 &62.99 &42.60 &79.08 \\
&SLS &93.79 &12.74 &86.42 &113.85 &47.94 &48.89 &45.56 &67.66 \\
&ALS &93.90 &12.96 &86.20 &114.51 &50.58 &46.14 &45.78 &66.17 \\
&TTSO &93.63 &45.96 &50.91 &162.66 &57.94 &38.12 &104.55 &44.52 \\
\cdashlinelr{2-11}
&TTSO++ &93.36 &\textbf{47.61} &\textbf{49.00} &\textbf{163.92} &\textbf{58.05} &\textbf{37.82} &\textbf{105.47} &\textbf{43.41} \\

\midrule
\multirow{9}{*}{MR} &- &87.15 &5.16 &94.08 &94.13 &33.30 &61.79 &47.11 &77.94 \\
&PGD &87.71 &5.16 &94.12 &96.62 &35.08 &60.00 &48.89 &77.06 \\
&FreeLB &88.27 &4.50 &94.90 &97.93 &35.37 &59.94 &47.54 &77.42 \\
&TAVAT &\textbf{89.59} &6.10 &93.19 &102.61 &38.37 &57.17 &49.03 &75.18 \\
&Flooding-X &88.18 &4.22 &95.21 &90.28 &33.49 &62.02 &46.54 &78.61 \\
&SLS &87.99 &13.32 &84.86 &124.05 &43.71 &50.32 &49.95 &67.59 \\
&ALS &88.37 &11.82 &86.62 &124.74 &42.96 &51.38 &51.01 &69.00 \\
&TTSO &88.46 &41.09 &53.55 &\textbf{171.37} &50.38 &43.05 &112.69 &48.30 \\
\cdashlinelr{2-11}
&TTSO++ &88.37 &\textbf{41.93} &\textbf{52.55} &170.61 &\textbf{51.97} &\textbf{41.19} &\textbf{114.71} &\textbf{46.87} \\

\midrule
\multirow{9}{*}{MRPC} &- &86.96 &4.81 &94.47 &152.11 &12.23 &85.93 &101.07 &90.20 \\
&PGD &86.55 &4.58 &94.71 &161.32 &11.88 &86.27 &104.67 &90.49 \\
&FreeLB &86.67 &5.45 &93.71 &152.83 &10.32 &88.09 &102.72 &90.90 \\
&TAVAT &\textbf{87.54} &4.87 &94.44 &174.73 &15.54 &82.25 &112.24 &88.34 \\
&Flooding-X &87.07 &4.81 &94.47 &149.09 &12.64 &85.49 &100.20 &89.98 \\
&SLS &86.84 &5.33 &93.86 &178.08 &19.36 &77.70 &119.15 &85.78 \\
&ALS &86.49 &8.41 &90.28 &175.10 &15.94 &81.57 &106.76 &85.92 \\
&TTSO &85.74 &43.94 &48.75 &380.00 &52.46 &38.81 &254.36 &43.78 \\
\cdashlinelr{2-11}
&TTSO++ &85.57 &\textbf{44.64} &\textbf{47.83} &\textbf{385.33} &\textbf{52.64} &\textbf{38.48} &\textbf{257.86} &\textbf{43.16} \\

\midrule
\multirow{9}{*}{SciTail} &- &94.54 &44.97 &52.44 &105.04 &35.32 &62.64 &96.17 &57.54 \\
&PGD &94.97 &49.81 &47.55 &109.80 &38.85 &59.09 &98.06 &53.32 \\
&FreeLB &94.78 &50.66 &46.55 &109.44 &40.36 &57.42 &104.09 &51.98 \\
&TAVAT &\textbf{95.11} &50.80 &46.59 &111.04 &42.00 &55.84 &102.13 &51.22 \\
&Flooding-X &93.27 &48.92 &47.55 &108.44 &38.62 &58.60 &99.93 &53.08 \\
&SLS &94.36 &48.31 &48.80 &112.80 &40.59 &56.98 &100.81 &52.89 \\
&ALS &94.07 &47.74 &49.25 &109.48 &40.59 &56.85 &97.75 &53.05 \\
&TTSO &94.40 &54.37 &42.40 &\textbf{123.80} &52.63 &44.25 &169.03 &43.33 \\
\cdashlinelr{2-11}
&TTSO++ &94.31 &\textbf{55.83} &\textbf{40.80} &123.65 &\textbf{53.86} &\textbf{42.89} &\textbf{171.95} &\textbf{41.84} \\
\bottomrule
\end{tabular}
\caption{The experiment results of different defence methods using BGE-M3 model. The best performance is marked in \textbf{bold}.}\label{tab:appendix_bge_m3}
\end{table*}

While BERT, RoBERTa, and DeBERTa are the most commonly used encoder-based models in prior studies \cite{raina-etal-2024-extreme, zhao2024disentangled, b35be69309694ab1b32513fe4e679660}, we extend this evaluation by assessing adversarial robustness using a more recent state-of-the-art embedding-based model BGE-M3 \cite{chen-etal-2024-m3}. 
Results are summarised in Table~\ref{tab:appendix_bge_m3}. 
TTSO++ consistently achieves superior robustness performance, excelling in all metrics across all datasets and attack types. 

\subsubsection{Robustness in Commonsense Reasoning}

\begin{table}[!t]\centering
\scriptsize
\begin{tabular}{lrrrrrr}\toprule
\multirow{2}{*}{\textbf{Dataset}} &\multirow{2}{*}{\textbf{Defence}} &\multirow{2}{*}{\textbf{\textsc{Acc}}$\uparrow$} &\multicolumn{3}{c}{\textbf{TextFooler}} \\\cmidrule{4-6}
& & &\textbf{\textsc{Aua}}$\uparrow$ &\textbf{\textsc{Asr}}$\downarrow$ &\textbf{\textsc{AvgQ}}$\uparrow$ \\\midrule
\multirow{9}{*}{SIQA} &- &71.24 &57.98 &18.61 &16.15 \\
&PGD &71.24 &58.39 &18.03 &16.18 \\
&FreeLB &71.55 &59.01 &17.53 &16.15 \\
&TAVAT &71.19 &62.33 &12.44 &16.32 \\
&Flooding-X &70.37 &57.98 &17.60 &16.13 \\
&SLS &71.60 &59.77 &16.51 &16.23 \\
&ALS &72.16 &59.77 &17.16 &16.12 \\
&TTSO &71.08 &59.21 &16.70 &16.23 \\
\cdashlinelr{2-6}
&TTSO++ &\textbf{72.22} &\textbf{63.01} &\textbf{11.36} &\textbf{17.01} \\
\midrule
\multirow{9}{*}{CSQA} &- &59.87 &48.24 &19.43 &11.94 \\
&PGD &59.11 &48.01 &18.77 &11.92 \\
&FreeLB &60.77 &48.73 &19.81 &11.99 \\
&TAVAT &\textbf{60.81} &48.51 &20.22 &12.01 \\
&Flooding-X &58.64 &47.42 &19.13 &11.91 \\
&SLS &59.46 &48.48 &18.46 &12.04 \\
&ALS &58.39 &46.76 &19.92 &12.04 \\
&TTSO &59.91 &49.12 &18.01 &11.62 \\
\cdashlinelr{2-6}
&TTSO++ &58.94 &\textbf{50.89} &\textbf{13.65} &\textbf{12.88} \\
\bottomrule
\end{tabular}
\caption{The experiment results on the commonsense reasoning tasks (SIQA and CSQA). Following \citet{branco2021shortcutted}, we employed TextFooler \cite{jin2020bert} and evaluated adversarial performance with RoBERTa-base under the same experimental settings. The best performance is marked in \textbf{bold}.}\label{tab:commonsense_reasoning_roberta}
\end{table}

Table~\ref{tab:commonsense_reasoning_roberta} highlights the adversarial robustness performance of all baseline defence methods on commonsense reasoning tasks using RoBERTa-base. 
TTSO++ achieves the best overall performance, with the highest \textsc{Aua} and lowest \textsc{Asr} across both datasets, demonstrating its strong defence capabilities. 
Flooding-X, however, consistently underperforms, reaffirming its limitations in adversarial settings. 
Notably, token-level embedding perturbation methods such as PGD, FreeLB, and TAVAT exhibit marginal improvements over the baseline but fail to achieve robustness comparable to TTSO++.



\section{Discussion}

\subsection{Dynamic Confidence Adjustment}

From Table~\ref{tab:main_v1}, we observe that TTSO++ consistently outperforms TTSO across datasets and models in terms of all evaluation metrics. 
A key factor in this improvement lies in the nuanced difference between the temperature-scaling mechanisms of TTSO and TTSO++.
TTSO applies a uniform temperature ($T_{base} =10$) to all instances during training, ensuring equal smoothing of logits across the dataset. 
While this strategy offers simplicity and improves model calibration, it is inherently limited. 
A fixed temperature does not account for variations in the difficulty of individual examples. 
For easy-to-classify examples (where the model is naturally confident), applying a slightly higher temperature can unnecessarily dampen the predictions, leading to a loss of useful certainty. 
Conversely, for hard-to-classify examples (where the model should be uncertain) or adversarial instances, applying a fixed high temperature may not be enough to capture the complexity of the example, leading to insufficient adjustment of the logits. 
In contrast, TTSO++ incorporates entropy-based temperature scaling, where the temperature is dynamically adjusted for each input instance based on the model's certainty. 
This approach leverages entropy as a proxy for uncertainty. 
Higher entropy (low certainty) leads to a higher temperature, while lower entropy (high certainty) results in a lower temperature. 
This adaptive mechanism allows TTSO++ to tailor the level of smoothing to the specific demands of each input, striking a better balance between preserving confidence for easy examples and enhancing robustness for challenging ones. 
As a result, TTSO++ achieves superior performance, where the ability to dynamically handle uncertain inputs is critical.

The effectiveness of TTSO++ is particularly evident in commonsense reasoning tasks like SIQA and CSQA (Table~\ref{tab:commonsense_reasoning_roberta}). 
Here, TTSO++ demonstrates the highest \textsc{Aua} and lowest \textsc{Asr} across all models and datasets. 
The instance-wise temperature scaling provides the model with the flexibility to adapt to diverse question-answering scenarios, effectively mitigating the impact of adversarial attacks. 
TTSO++ sets a new benchmark, offering superior adversarial robustness and generalisability across datasets and tasks.

\subsection{High Temperature Training}

\begin{figure}
    \centering
    \includegraphics[width=1.0\linewidth]{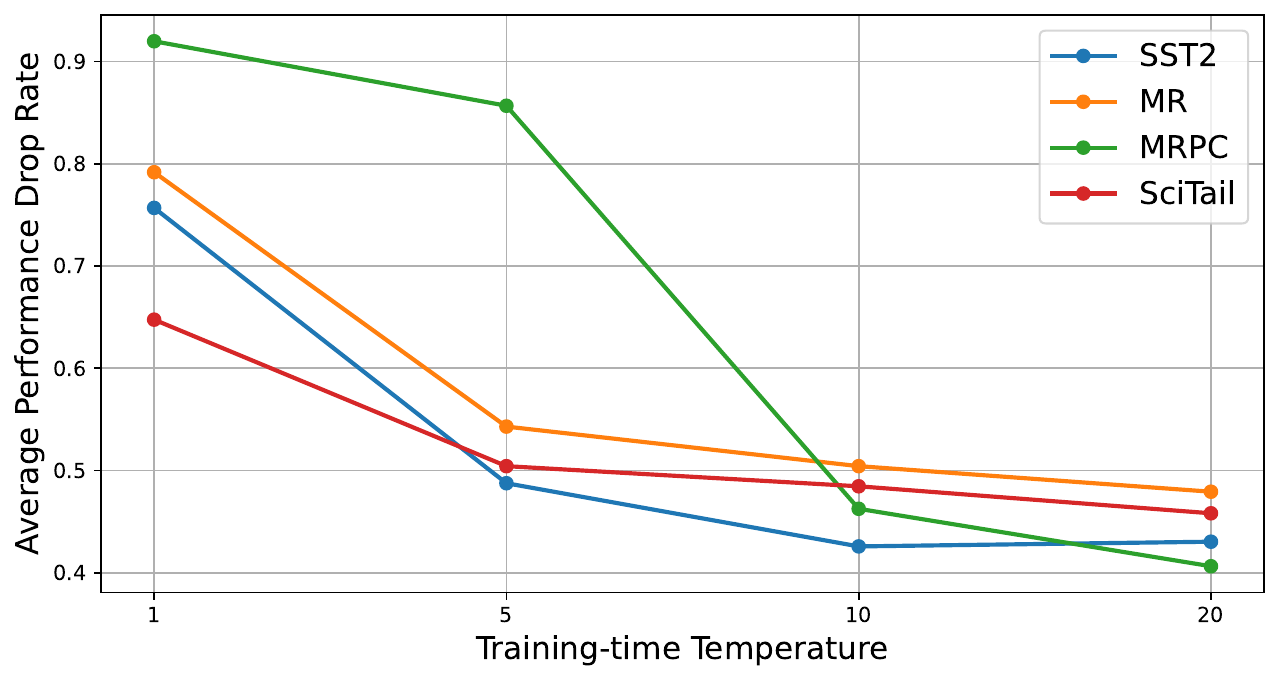}
    \caption{Average performance drop rate (\textsc{Apdr}) across two attackers using TTSO++ with RoBERTa-base as training temperature varies.}
    \label{fig:apdr}
\end{figure}

While tuning $T_{base}$ could potentially enhance performance against adversarial attacks, we opted for a fixed temperature to ensure consistency and simplicity in our experimental setup. 
The choice of 10 as the fixed temperature was empirically validated across a range of NLP tasks and demonstrated robust performance across clean and adversarial examples. 
By using a fixed temperature, we reduce the need for extensive hyper-parameter tuning, which can introduce additional computational overhead and potential overfitting to specific datasets or adversarial attacks. 

Figure~\ref{fig:apdr} presents the changes in before- and after-attack accuracy of a model trained with the standard objective and various base temperatures ($T_{base}$, as described in \S\ref{section:adversarial_defence}) during training.
While similar trends were observed across all models, we present the results specifically for RoBERTa-base in this section. 
The results indicate that higher temperatures during training generally enhance robustness against adversarial attacks. 
To quantify this, we use the average performance drop rate (\textsc{Apdr}) \cite{zhu2023promptbench}, which averages the performance drop rate 
\begin{equation}
    \textsc{Pdr} = 1 - \frac{\sum_{(x;y) \in \mathcal{D}} { \mathcal{M} [ f_{\theta}(A(x)), y]}}{ \sum_{(x;y) \in \mathcal{D}} {\mathcal{M} [f_{\theta}(x), y]}}
\end{equation}
across different adversarial attacks, where $A$ is the adversarial attack applied to input text $x$, $\mathcal{M}[\cdot]$ is the evaluation function, and $f_{\theta}(\cdot)$ is the network. 
For classification task, $\mathcal{M}[\cdot]$ is the indicator function $\mathbbm{1} [\hat{y}, y]$ which equals to 1 when $\hat{y} = y$, and 0 otherwise. 
Notably, on SST2 dataset, we observe a slight increase in \textsc{Apdr} when the training temperature is set to 20. 
This suggests that excessively high temperatures may overly smooth the predicted probability distribution, making it harder for the model to effectively learn from the training data.

\subsection{Runtime Analysis}

\begin{table}[!t]\centering
\scriptsize
\begin{tabular}{lrrrr}\toprule
\textbf{Defence} &\textbf{SST2} &\textbf{MRPC} &\textbf{SIQA} \\\midrule
- &1 &1 &1 \\
\midrule
PGD &x6.3 &x10.2 &x4.1 \\
FreeLB &x2.6 &x3.1 &x2.2 \\
TAVAT &x4.2 &x3.9 &x2.6 \\
\midrule
Flooding-X &x1.2 &x1.3 &x1.1 \\
SLS &x1.1 &x1.1 &x1.1 \\
ALS &x1.2 &x1.2 &x1.1 \\
TTSO &x1.1 &x1.1 &x1.0 \\
\midrule
TTSO++ &x1.1 &x1.1 &x1.1 \\
\bottomrule
\end{tabular}
\caption{Runtime comparison of training RoBERTa-base using different adversarial defence methods.}\label{tab:runtime}
\end{table}

Table~\ref{tab:runtime} presents the runtime comparison of training the RoBERTa-base model on the SST2, MRPC, and SIQA datasets using different adversarial defence methods. The baseline model (no defence) has a normalised runtime of 1 across all datasets, serving as the standard for comparison.

Adversarial training-based methods such as PGD, FreeLB, and TAVAT introduce significant runtime overhead due to the inclusion of inner maximisation steps. PGD, which requires multiple gradient updates per iteration to approximate adversarial perturbations, is the most computationally expensive, resulting in training times that are x6.3, x10.2, and x4.1 longer for SST2, MRPC, and SIQA, respectively. 
Although adversarial training-based methods offer slight improvements in adversarial robustness, the runtime cost makes them impractical for large-scale training tasks. 
In contrast, regularisation-based methods like Flooding-X, SLS, ALS, TTSO, and TTSO++ impose minimal runtime overhead, with training times ranging from x1.1 to x1.3 across all datasets. 
These methods are particularly appealing for large-scale scenarios, as they do not involve the computationally expensive inner maximisation step. 
Among these, TTSO++ stands out by combining strong adversarial robustness with a minimal runtime impact. 
Its entropy-based temperature scaling mechanism effectively adjusts model predictions without requiring extensive computational resources, making it an ideal defence for both efficiency and robustness.




\section{Conclusion}

In this work, we investigated adversarial defence techniques that are broadly applicable across diverse NLP tasks, focusing on synonym-agnostic and structure-free approaches. 
By establishing a comprehensive benchmark, we evaluated state-of-the-art adversarial defence strategies developed prior to 2024, extending the evaluation beyond traditional text classification to encompass single-sentence classification, similarity and paraphrase identification, natural language inference, and commonsense reasoning tasks.



Our systematic exploration of regularisation-based methods revealed valuable insights into their potential for textual adversarial defence. 
Based on these findings, we proposed TTSO++, a simple yet effective variant of temperature scaling that leverages entropy-based adjustments during training. 
TTSO++ achieves state-of-the-art robustness under adversarial attacks while maintaining strong performance on clean examples. 
Its minimal computational overhead makes it highly practical for real-world applications. 
By extending adversarial evaluation to a broader spectrum of NLP tasks, we aim to inspire the development of more flexible, generalisable, and efficient defence mechanisms. 
We believe this study provides a robust foundation for future research, bridging the gap between task-specific defences and universally applicable solutions for adversarial robustness in NLP. 

\section*{Limitations}

Our study presents empirical results using state-of-the-art encoder-based Transformer models, which are widely regarded as the most appropriate for classification-based NLP tasks \cite{raina-etal-2024-extreme, zhao2024disentangled}. 
However, the rapidly growing field of LLMs opens new avenues for exploration. 
Future work could examine the susceptibility of decoder-based LLMs to adversarial attacks and evaluate the performance of the defence methods discussed in this paper in such settings. 
Additionally, while our research focuses on defence methods that can be uniformly applied across all benchmark datasets, it remains unexplored whether more specialised techniques, such as contrastive-based methods \cite{pan2022improved, 10.3233/JIFS-230787} or prompt-based methods \cite{xu2024textit, yang2024prompt}, could be adapted to provide universal adversarial defences. 
Investigating these methods' applicability to a broader range of tasks could further enhance the scope of adversarial robustness research.
Finally, we proposed TTSO++ as an improvement over the fixed-temperature TTSO method by introducing entropy-based temperature scaling. 
While TTSO++ demonstrates significant advancements, further optimisation of temperature scaling strategies could yield additional improvements. 
For example, dynamically adjusting the temperature based on training progression (e.g., curriculum-based or confidence-based scaling) may better align with the evolving complexity of the task during training. 
Future research could explore these methods to develop more adaptive and effective defences.

\section*{Acknowledgments}

This work was supported by the Innovate UK Knowledge Transfer Partnership (KTP) grant (Grant Number: 13320). We thank the University of Manchester for providing the computing resources required to conduct the experiments.


\bibliography{custom}

\appendix

\section{Generative LLMs}
\label{appendix:generative_llms}

With the advent of powerful generative large language models (LLMs), such as ChatGPT \cite{chatgpt}, their usage has become increasingly widespread. However, similar to recent studies \cite{zhong2023can, raina-etal-2024-extreme, periti2024chat}, we find that these popular generative LLMs are not suitable for inclusion in our benchmark for several key reasons. A comparative analysis of their performance with state-of-the-art generative LLMs, using 0-shot and few-shot prompting, is presented in Table~\ref{tab:generative_llms} for some datasets considered in this paper. 
First, fine-tuned encoder-based models (e.g., BERT-based models) continue to demonstrate competitive, if not superior, performance on each task, which has led to their extensive adoption in many industry applications. These models are not only lightweight (possessing far fewer parameters compared to generative LLMs) but also cost-effective while achieving strong performance across a wide range of tasks. 
Second, the use of generative LLMs complicates the standardisation of input-output formats across diverse tasks and datasets, which introduces potential bias into the evaluation process \cite{liu2023pre}. This challenge makes it difficult to ensure reproducibility and to facilitate fair comparisons across different research contexts \cite{hayase2024query, rathje2024gpt}. Moreover, disentangling a model's intrinsic performance from artifacts introduced by the prompting strategy is non-trivial, as model outcomes are influenced by both the design of the prompts and the generated responses \cite{gao-etal-2021-making, liu2023pre}.
Finally, the adversarial attack and defence literature, which forms the basis of our contributions, predominantly focuses on encoder-based models. Aligning our experimental setup with this body of work enables us to build on existing attack and defence mechanisms.

In Table~\ref{tab:generative_llms}, we provide a comparative analysis of the performance of several state-of-the-art generative LLMs, utilising both zero-shot and five-shot prompting. 
Additionally, we present performance comparisons from \citet{zhong2023chatgptunderstandtoocomparative} and \citet{raina-etal-2024-extreme}, and evaluate encoder-based models (BERT, RoBERTa, and DeBERTa) alongside generative LLMs, including Phi3 \cite{abdin2024phi} and Llama3 \cite{dubey2024llama}, on some of the datasets covered in this work.

\begin{table}[!t]\centering
\resizebox{\linewidth}{!}{
\begin{tabular}{lccccc}\toprule
\textbf{Model} &\textbf{Params} &\textbf{SST2 (\%)} &\textbf{MR (\%)} &\textbf{MRPC (\%)} \\\midrule
Mistral-7B (0-shot)$^{\dag}$ &7B &- &86.47 &67.15 \\
Mistral-7B (5-shot)$^{\dag}$ &7B &- &88.92 &76.21 \\
\midrule
ChatGPT-3.5 (0-shot)$^{\ddag}$ &- &92.00 &- &66.00 \\
ChatGPT-3.5 (1-shot)$^{\ddag}$ &- &96.00 &- &66.00 \\
ChatGPT-3.5 (5-shot)$^{\ddag}$ &- &98.00 &- &76.00 \\
ChatGPT-3.5 (0-shot CoT)$^{\ddag}$ &- &96.00 &- &78.00 \\
\midrule
BERT-base &110M &91.54 &85.55 &84.40 \\
RoBERTa-base &110M &93.79 &88.65 &86.96 \\
DeBERTa-base &110M &95.22 &90.71 &87.94 \\
Phi3-3.8B (0-shot) &3.8B &85.93 &81.57 &74.01 \\
Phi3-3.8B (0-shot CoT) &3.8B &87.19 &83.44 &74.29 \\
Llama3-8B (0-shot) &8B &89.46 &83.80 &76.90 \\
Llama3-8B (0-shot CoT) &8B &90.12 &84.13 &78.49 \\
\bottomrule
\end{tabular}
}
\caption{Comparison of model performance with popular generative LLMs. $\dag$ Figures given in \citet{raina-etal-2024-extreme}. $\ddag$ Figures given in \citet{zhong2023chatgptunderstandtoocomparative}.}\label{tab:generative_llms}
\end{table}

\section{Detailed Performance Breakdown}
\label{appendix:performance_breakdown}

In this section, we provide the detailed breakdown of performances for the different Transformer encoders across each dataset: Table~\ref{tab:appendix_bert} for BERT-base model, Table~\ref{tab:appendix_roberta} for RoBERTa-base model, and Table~\ref{tab:appendix_deberta} for DeBERTa-base model. 
Each Table presents the adversarial robustness performance trained with different defence methods. 

\begin{table*}[!t]\centering
\scriptsize
\begin{tabular}{lrrrrrrrrrr}\toprule
\multirow{2}{*}{\textbf{Dataset}} &\multirow{2}{*}{\textbf{Defence}} &\multirow{2}{*}{\textbf{\textsc{Acc}$\uparrow$}} &\multicolumn{3}{c}{\textbf{TextFooler (\%)}} &\multicolumn{3}{c}{\textbf{TextBugger (\%)}} &\multirow{2}{*}{\textbf{\textsc{Apdr}$\downarrow$}} \\\cmidrule{4-9}
& & &\textbf{\textsc{Aua}$\uparrow$} &\textbf{\textsc{Asr}$\downarrow$} &\textbf{\textsc{AvgQ}$\uparrow$} &\textbf{\textsc{Aua}$\uparrow$} &\textbf{\textsc{Asr}$\downarrow$} &\textbf{\textsc{AvgQ}$\uparrow$} & \\\midrule
\multirow{8}{*}{SST2} 
&- &91.54 &6.59 &92.80 &89.53 &28.08 &69.35 &41.30 &81.06 \\
&PGD &\textbf{92.64} &8.73 &90.57 &98.54 &34.27 &63.01 &42.74 &76.79 \\
&FreeLB &91.98 &8.57 &90.69 &98.95 &31.80 &65.43 &42.35 &78.06 \\
&TAVAT &\textbf{92.64} &10.71 &88.44 &103.78 &34.32 &62.95 &43.21 &75.70 \\
&Flooding-X &89.84 &11.64 &87.04 &95.38 &30.37 &66.20 &43.08 &76.62 \\
&SLS &91.76 &12.25 &86.65 &108.85 &39.21 &57.27 &44.74 &71.96 \\
&ALS &91.21 &15.54 &82.96 &110.51 &39.37 &56.83 &46.48 &69.90 \\
&TTSO &91.71 &\textbf{41.63} &\textbf{54.61} &\textbf{148.27} &\textbf{50.85} &\textbf{44.55} &\textbf{95.83} &\textbf{49.58} \\
\midrule
\multirow{8}{*}{MR} 
&- &85.55 &3.94 &95.39 &82.70 &22.80 &73.36 &42.85 &84.37 \\
&PGD &85.93 &12.85 &85.04 &115.93 &35.46 &58.73 &49.77 &71.89 \\
&FreeLB &86.30 &6.66 &92.28 &99.94 &28.71 &66.74 &45.51 &79.51 \\
&TAVAT &86.02 &8.26 &90.40 &107.21 &30.11 &64.99 &47.04 &77.70 \\
&Flooding-X &85.55 &5.44 &93.64 &91.97 &27.02 &68.42 &44.77 &81.03 \\
&SLS &\textbf{86.59} &13.60 &84.29 &118.28 &36.49 &57.85 &47.95 &71.08 \\
&ALS &85.83 &11.07 &87.10 &112.63 &33.77 &60.66 &49.31 &73.88 \\
&TTSO &86.02 &\textbf{35.27} &\textbf{59.00} &\textbf{157.94} &\textbf{43.06} &\textbf{49.95} &\textbf{103.85} &\textbf{54.47} \\
\midrule
\multirow{8}{*}{MRPC} 
&- &84.40 &2.32 &97.25 &124.00 &3.25 &96.15 &72.84 &96.70 \\
&PGD &84.06 &9.86 &88.28 &205.38 &11.25 &86.62 &101.98 &87.44 \\
&FreeLB &\textbf{85.45} &11.48 &86.57 &212.41 &11.65 &86.36 &107.19 &86.47 \\
&TAVAT &84.29 &8.70 &89.68 &229.16 &10.43 &87.62 &106.60 &88.65 \\
&Flooding-X &82.67 &7.48 &90.95 &151.69 &7.88 &90.46 &86.56 &90.71 \\
&SLS &84.52 &6.32 &92.52 &170.88 &7.36 &91.29 &88.79 &91.91 \\
&ALS &82.96 &5.97 &92.80 &168.67 &7.83 &90.57 &93.89 &91.68 \\
&TTSO &83.77 &\textbf{41.62} &\textbf{50.31} &\textbf{370.89} &\textbf{39.71} &\textbf{52.60} &\textbf{220.98} &\textbf{51.46} \\
\midrule
\multirow{8}{*}{SciTail} 
&- &92.80 &44.45 &52.10 &106.17 &32.22 &65.28 &95.52 &58.69 \\
&PGD &93.09 &50.19 &46.08 &111.88 &32.69 &64.88 &95.07 &55.48 \\
&FreeLB &\textbf{93.60} &47.04 &49.75 &109.00 &32.60 &65.18 &96.17 &57.46 \\
&TAVAT &92.29 &\textbf{52.21} &\textbf{43.43} &115.19 &30.48 &66.97 &98.18 &55.20 \\
&Flooding-X &91.58 &49.62 &45.81 &111.23 &35.28 &61.48 &102.06 &53.65 \\
&SLS &92.33 &48.02 &47.99 &110.77 &34.85 &62.25 &94.86 &55.12 \\
&ALS &92.57 &50.80 &45.12 &112.64 &33.44 &63.87 &97.90 &54.50 \\
&TTSO &92.33 &52.02 &43.66 &\textbf{123.38} &\textbf{48.21} &\textbf{47.78} &\textbf{157.87} &\textbf{45.72} \\
\bottomrule
\end{tabular}
\caption{The experiment results of different defence methods using BERT-base model.}\label{tab:appendix_bert}
\end{table*}

\begin{table*}[!t]\centering
\scriptsize
\begin{tabular}{lrrrrrrrrrr}\toprule
\multirow{2}{*}{\textbf{Dataset}} &\multirow{2}{*}{\textbf{Defence}} &\multirow{2}{*}{\textbf{\textsc{Acc}$\uparrow$}} &\multicolumn{3}{c}{\textbf{TextFooler}} &\multicolumn{3}{c}{\textbf{TextBugger}} &\multirow{2}{*}{\textbf{\textsc{Apdr}$\downarrow$}} \\\cmidrule{4-9}
& & &\textbf{\textsc{Aua}$\uparrow$} &\textbf{\textsc{Asr}$\downarrow$} &\textbf{\textsc{AvgQ}$\uparrow$} &\textbf{\textsc{Aua}$\uparrow$} &\textbf{\textsc{Asr}$\downarrow$} &\textbf{\textsc{AvgQ}$\uparrow$} & \\\midrule
\multirow{8}{*}{SST2} 
&- &93.79 &7.80 &91.69 &89.15 &31.52 &66.39 &43.31 &79.04 \\
&PGD &94.29 &8.79 &90.68 &97.68 &34.21 &63.72 &44.28 &77.20 \\
&FreeLB &95.11 &6.43 &93.24 &94.88 &36.90 &61.20 &44.23 &77.22 \\
&TAVAT &\textbf{95.17} &9.23 &90.31 &100.21 &37.40 &60.70 &44.25 &75.50 \\
&Flooding-X &94.95 &5.44 &94.27 &85.27 &29.65 &68.77 &41.99 &81.52 \\
&SLS &94.34 &11.64 &87.66 &111.72 &44.81 &52.50 &45.69 &70.08 \\
&ALS &94.51 &23.50 &75.13 &130.85 &52.50 &44.45 &49.03 &59.79 \\
&TTSO &94.67 &\textbf{45.63} &\textbf{51.80} &\textbf{159.93} &\textbf{56.84} &\textbf{39.97} &\textbf{101.98} &\textbf{45.88} \\
\midrule
\multirow{8}{*}{MR} 
&- &88.65 &6.75 &92.38 &101.06 &31.80 &64.13 &48.91 &78.26 \\
&PGD &87.90 &6.29 &92.85 &102.56 &34.33 &60.94 &49.29 &76.89 \\
&FreeLB &\textbf{89.12} &5.53 &93.79 &98.94 &33.58 &62.32 &47.84 &78.06 \\
&TAVAT &87.99 &6.19 &92.96 &102.04 &32.46 &63.11 &48.46 &78.04 \\
&Flooding-X &88.74 &7.79 &91.23 &99.61 &31.05 &65.01 &46.73 &78.12 \\
&SLS &88.09 &20.83 &76.36 &136.77 &42.59 &51.65 &53.54 &64.00 \\
&ALS &87.90 &17.07 &80.58 &130.26 &43.34 &50.69 &53.59 &65.64 \\
&TTSO &87.71 &\textbf{42.40} &\textbf{51.66} &\textbf{171.72} &\textbf{51.97} &\textbf{40.75} &\textbf{114.56} &\textbf{46.20} \\
\midrule
\multirow{8}{*}{MRPC} 
&- &86.96 &6.09 &93.00 &163.57 &9.57 &89.00 &96.71 &91.00 \\
&PGD &87.48 &5.28 &93.97 &180.35 &11.54 &86.81 &102.54 &90.39 \\
&FreeLB &87.54 &6.38 &92.72 &191.58 &11.71 &86.62 &105.26 &89.67 \\
&TAVAT &\textbf{87.59} &9.97 &88.62 &212.02 &15.07 &82.79 &108.67 &85.71 \\
&Flooding-X &87.48 &4.75 &94.57 &173.10 &8.41 &90.39 &101.31 &92.48 \\
&SLS &86.84 &9.22 &89.39 &208.54 &12.35 &85.78 &111.62 &87.58 \\
&ALS &86.03 &10.61 &87.67 &220.83 &13.28 &84.57 &108.99 &86.12 \\
&TTSO &86.78 &\textbf{46.38} &\textbf{46.56} &\textbf{389.29} &\textbf{41.91} &\textbf{51.70} &\textbf{226.51} &\textbf{49.13} \\
\midrule
\multirow{8}{*}{SciTail} 
&- &93.60 &42.80 &54.27 &101.92 &31.70 &66.13 &92.84 &60.20 \\
&PGD &93.09 &43.27 &53.51 &104.29 &31.42 &66.25 &95.29 &59.88 \\
&FreeLB &\textbf{93.79} &44.21 &52.86 &104.06 &31.51 &66.40 &96.19 &59.63 \\
&TAVAT &\textbf{93.79} &46.38 &50.55 &106.96 &34.71 &62.99 &98.50 &56.77 \\
&Flooding-X &92.43 &43.32 &53.13 &104.57 &29.02 &68.60 &92.31 &60.87 \\
&SLS &93.60 &45.58 &51.31 &106.60 &35.79 &61.76 &103.88 &56.53 \\
&ALS &92.90 &44.17 &52.46 &105.72 &34.48 &62.89 &93.94 &57.67 \\
&TTSO &92.52 &\textbf{51.27} &\textbf{44.59} &\textbf{121.66} &\textbf{47.22} &\textbf{48.96} &\textbf{158.72} &\textbf{46.77} \\
\bottomrule
\end{tabular}
\caption{The experiment results of different defence methods using RoBERTa-base model.}\label{tab:appendix_roberta}
\end{table*}

\begin{table*}[!t]\centering
\scriptsize
\begin{tabular}{lrrrrrrrrrr}\toprule
\multirow{2}{*}{\textbf{Dataset}} &\multirow{2}{*}{\textbf{Defence}} &\multirow{2}{*}{\textbf{\textsc{Acc}$\uparrow$}} &\multicolumn{3}{c}{\textbf{TextFooler}} &\multicolumn{3}{c}{\textbf{TextBugger (\%)}} &\multirow{2}{*}{\textbf{\textsc{Apdr}$\downarrow$ (\%)}} \\\cmidrule{4-9}
& & &\textbf{\textsc{Aua}$\uparrow$} &\textbf{\textsc{Asr}$\downarrow$} &\textbf{\textsc{AvgQ}$\uparrow$} &\textbf{\textsc{Aua}$\uparrow$} &\textbf{\textsc{Asr}$\downarrow$} &\textbf{\textsc{AvgQ}$\uparrow$} & \\\midrule
\multirow{8}{*}{SST2} &- &95.22 &8.57 &91.00 &96.80 &39.65 &58.36 &44.45 &74.68 \\
&PGD &94.67 &8.73 &90.78 &96.96 &38.88 &58.93 &44.15 &74.85 \\
&FreeLB &95.22 &10.32 &89.16 &108.13 &44.10 &53.69 &45.09 &71.42 \\
&TAVAT &\textbf{95.83} &11.04 &88.48 &111.03 &47.01 &50.95 &46.74 &69.71 \\
&Flooding-X &95.22 &7.36 &92.27 &92.93 &31.74 &66.67 &42.09 &79.47 \\
&SLS &95.22 &22.24 &76.64 &130.72 &54.48 &42.79 &47.70 &59.71 \\
&ALS &95.55 &15.76 &83.51 &120.09 &52.22 &45.34 &49.00 &64.43 \\
&TTSO &95.66 &\textbf{55.02} &\textbf{42.48} &\textbf{173.84} &\textbf{65.84} &\textbf{31.17} &\textbf{109.65} &\textbf{36.83} \\
\midrule
\multirow{8}{*}{MR} &- &90.71 &9.94 &89.04 &101.00 &34.24 &62.25 &47.71 &75.65 \\
&PGD &89.21 &5.35 &94.01 &96.21 &30.11 &66.25 &47.00 &80.13 \\
&FreeLB &91.18 &8.63 &90.53 &105.07 &34.80 &61.83 &48.37 &76.18 \\
&TAVAT &90.90 &10.60 &88.34 &110.67 &38.37 &57.79 &50.01 &73.06 \\
&Flooding-X &\textbf{91.37} &6.66 &92.71 &98.27 &35.46 &61.19 &47.03 &76.95 \\
&SLS &90.62 &17.45 &80.75 &135.75 &45.31 &50.00 &52.15 &65.37 \\
&ALS &89.96 &17.92 &80.08 &132.54 &46.53 &48.28 &53.05 &64.18 \\
&TTSO &90.34 &\textbf{47.84} &\textbf{47.04} &\textbf{179.78} &\textbf{56.75} &\textbf{37.18} &\textbf{117.93} &\textbf{42.11} \\
\midrule
\multirow{8}{*}{MRPC} &- &87.94 &2.96 &96.64 &155.80 &9.86 &88.79 &97.04 &92.71 \\
&PGD &87.83 &3.71 &95.78 &170.55 &10.55 &87.99 &98.00 &91.88 \\
&FreeLB &86.43 &8.58 &90.07 &186.01 &15.07 &82.56 &107.91 &86.32 \\
&TAVAT &88.06 &6.61 &92.50 &194.10 &16.23 &81.57 &110.88 &87.03 \\
&Flooding-X &88.35 &3.30 &96.26 &156.63 &10.20 &88.45 &97.65 &92.36 \\
&SLS &\textbf{88.46} &8.58 &90.30 &158.34 &12.87 &85.45 &101.25 &87.88 \\
&ALS &\textbf{88.46} &6.14 &93.05 &182.10 &17.45 &80.28 &111.49 &86.67 \\
&TTSO &87.54 &\textbf{50.78} &\textbf{41.99} &\textbf{394.67} &\textbf{54.38} &\textbf{37.88} &\textbf{253.36} &\textbf{39.94} \\
\midrule
\multirow{8}{*}{SciTail} &- &95.53 &47.04 &50.76 &107.60 &33.82 &64.60 &85.38 &57.68 \\
&PGD &94.36 &43.32 &54.09 &106.32 &30.39 &67.80 &89.40 &60.94 \\
&FreeLB &95.67 &47.04 &50.84 &107.88 &33.68 &64.80 &90.24 &57.81 \\
&TAVAT &\textbf{96.52} &50.19 &48.00 &112.98 &39.98 &58.58 &98.80 &53.29 \\
&Flooding-X &94.73 &46.28 &51.14 &106.81 &32.22 &65.99 &85.41 &58.57 \\
&SLS &95.67 &48.64 &49.16 &110.09 &35.04 &63.37 &92.54 &56.27 \\
&ALS &95.16 &50.85 &46.56 &116.84 &43.09 &54.72 &103.17 &50.64 \\
&TTSO &95.63 &\textbf{55.13} &\textbf{42.35} &\textbf{126.89} &\textbf{50.05} &\textbf{47.66} &\textbf{164.70} &\textbf{45.01} \\
\bottomrule
\end{tabular}
\caption{The experiment results of different defence methods using DeBERTa-base model.}\label{tab:appendix_deberta}
\end{table*}

\end{document}